\documentclass[conference]{IEEEtran}
\IEEEoverridecommandlockouts
\usepackage{cite}
\usepackage{amsmath,amssymb,amsfonts}
\usepackage{algorithmic}
\usepackage{graphicx}
\usepackage{textcomp}
\usepackage{xcolor}
\usepackage{subcaption}
\usepackage{algorithm}
\usepackage{algorithmic}

\def\BibTeX{{\rm B\kern-.05em{\sc i\kern-.025em b}\kern-.08em
    T\kern-.1667em\lower.7ex\hbox{E}\kern-.125emX}}
\begin{document}

\title{Finding Needles in Haystack: Formal Generative Models for Efficient Massive Parallel Simulations\\
{\footnotesize }
\thanks{This work is part of the project ``KImaDiZ", supported by the German Aerospace Center (DLR) with funds of the German Federal Ministry of Economics and Technology (BMWi), support code 50 RA 1934.}
}

\author{\IEEEauthorblockN{Osama Maqbool}
\IEEEauthorblockA{\textit{Institute of Man-Machine-Interaction} \\
\textit{RWTH Aachen University}\\
Aachen, Germany \\
maqbool@mmi.rwth-aachen.de}
\and
\IEEEauthorblockN{Jürgen Roßmann}
\IEEEauthorblockA{\textit{Institute for Man-Machine Interaction} \\
\textit{RWTH Aachen University}\\
Aachen, Germany \\
rossmann@mmi.rwth-aachen.de}
}

\maketitle

\begin{abstract}
The increase in complexity of autonomous systems is accompanied by a need of data-driven development and validation strategies. Advances in computer graphics and cloud clusters have opened the way to massive parallel high fidelity simulations to qualitatively address the large number of operational scenarios. However, exploration of all possible scenarios is still prohibitively expensive and outcomes of scenarios are generally unknown apriori. To this end, the authors propose a method based on bayesian optimization to efficiently learn generative models on scenarios that would deliver desired outcomes (e.g. collisions) with high probability. The methodology is integrated in an end-to-end framework, which uses the OpenSCENARIO standard to describe scenarios, and deploys highly configurable digital twins of the scenario participants on a Virtual Test Bed cluster. 
\end{abstract}

\begin{IEEEkeywords}
massive parallel simulations, bayesian optimization, virtual test beds, experimentable digital twins
\end{IEEEkeywords}

\section{Introduction}
The advent of intelligent vehicles has brought with it increasing levels of system complexity. Vehicles with machine learning components additionally incorporate black-boxes and uncertainty within the system, transforming an already difficult problem to a non-deterministic one. These systems therefore require data-driven strategies in addition to classical approaches to deliver statistical metrics on safety and reliability. Simulations offer a natural supplement to real-world tests, allowing reproduction of expensive or dangerous scenarios virtually and being scaled as needed. Advancements in computer graphics has made high fidelity simulations possible, which are especially beneficial for generating large volumes of realistic sensor data required for machine learning based perception systems. More recently, the availability of cloud computing resources, e.g. Microsoft Azure, has offset the procurement effort of on-premise compute clusters enabling large-scale parallel simulations for a wider community.

\begin{figure}
	\centering
	\includegraphics[width=0.4\textwidth]{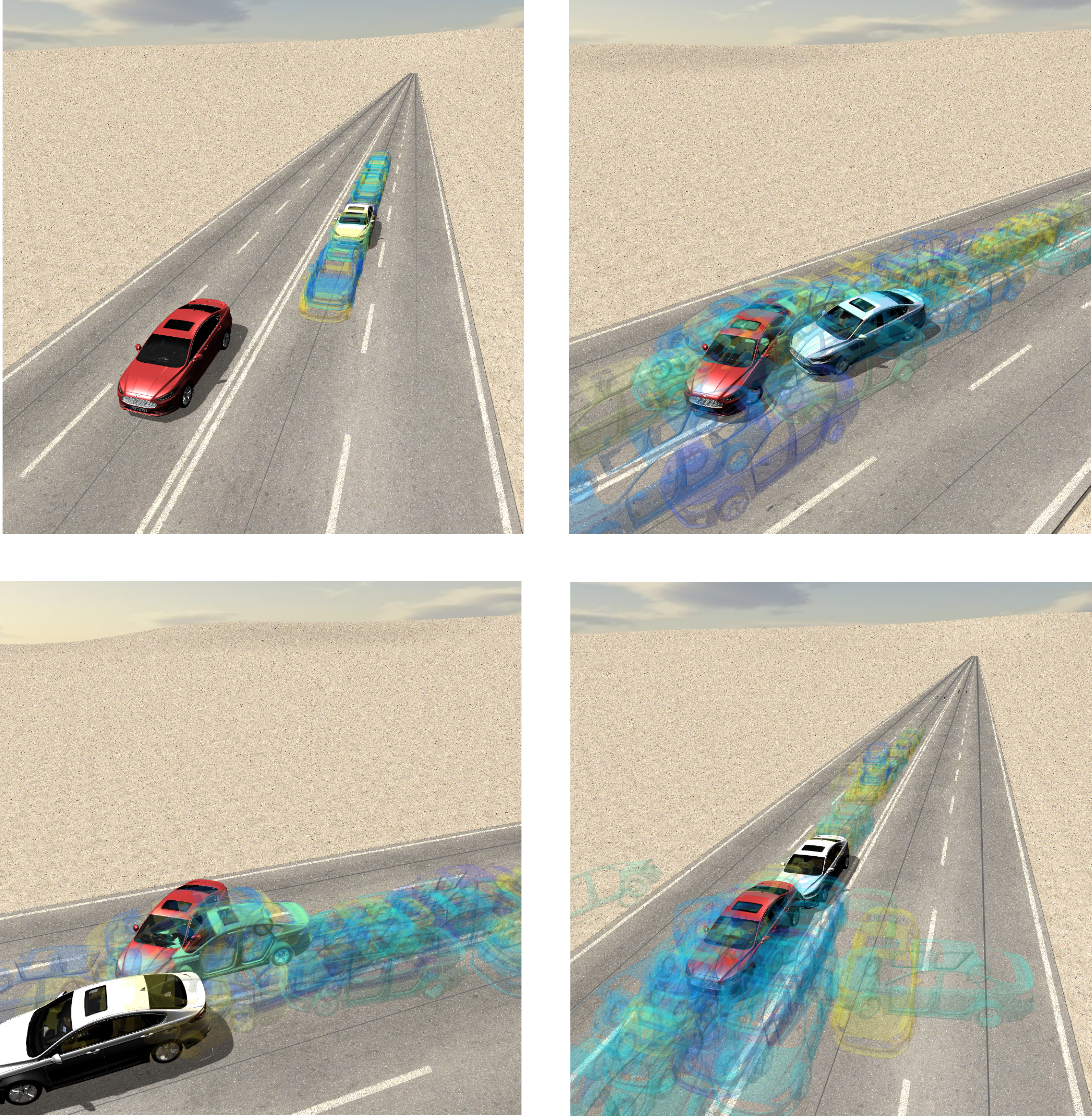}%
	\caption{Visualizations of \textit{cut-in-from-left} simulation. Transparent ``ghosts" illustrate parallel variants at the same time-point.}
	\label{fig:cutin_examples}
\end{figure}

Nevertheless, the space of all possible operational scenarios remains prohibitively large, and usage of on-premise or cloud-based compute clusters is additionally a cost-incurring process. This warrants a motivation to only simulate scenarios that would be meaningful to the actual development process. For instance, the training process of a machine learning based accident-prevention system would require simulations either violating a safety metric or close to the point of violation. Assuming knowledge of the system, this can be achieved by constraints on the type of scenarios, i.e. inputs to the simulation \cite{fremont2019scenic} \cite{MaqboolR22}. Intelligent vehicular systems, however are black-box systems and outcomes of scenarios are rarely known \textit{apriori}. Although extensive literature exists on adversarial validation of systems \cite{9090897}, i.e. iterative search within the scenario space to yield failure outcomes, these are generally falsification approaches and do not suffice to generate the large quantity of scenarios required for comprehensive coverage of system behaviors.

This paper proposes an efficient and flexible methodology to learn generative models over the scenario space with respect to given outcomes. Specifying outcome metrics as cost functions on simulation traces, we employ bayesian optimization \cite{jones1998efficient} to learn a surrogate function as belief over the cost function with reasonable confidence around minima. The surrogate is then used for fitting generative models. The methodology is formalized in an end-to-end framework which uses the OpenSCENARIO standard \cite{OpenScenario} to specify scenarios and brings them to life by constructing digital twins of scenario participants in a  virtual test bed (VTB) cluster. The setup is tested on a highway scenario, some examples of which are illustrated in Fig. \ref{fig:cutin_examples}.

The rest of the paper is structured as follows: Sec. \ref{sec:related_work} provides a survey of related work followed by a formal problem definition in Sec. \ref{sec:problem} and introduction to theoretical methods in Sec. \ref{sec:Background}. Sec. \ref{sec:Framework} explains the full architecture that employs the proposed methodology, and Sec. \ref{sec:application} applies the methodology to an example application, finally followed by the conclusion.

\section{Related Work}\label{sec:related_work}
The scenario-based architecture used by the authors is based on the PEGASUS project \cite{winner2019pegasus}, which aimed towards standardized safety qualification of autonomous vehicles and has spawned various works for formal scenario description \cite{kramer2020identification}\cite{menzel2018scenarios} and derivations of critical scenarios via data- \cite{putz2017system} or knowledge-driven \cite{bagschik2018wissensbasierte} methods. Scenic \cite{fremont2019scenic} provides another formal description language for scenarios, which is similarly used for formal safety qualifications \cite{fremont2020formal}.

Stepping outside the formalized scenario domain, many approaches tackle the validation problem of systems with machine learning (ML) components via falsification - typically optimization-based search policies to find counter-examples for systems. Among them are image generation methods, such as domain randomization in \cite{8794443} and \cite{khirodkar2018vadra}, or similarly executed point clouds generation approaches as in \cite{abdelfattah2021towards}. Such approaches attack the object-detection or planning ML directly, but the exploited parameters lack semantic information about the scenario and do not offer a holistic interpretation with respect to the system. To this end, VERIFAI \cite{dreossi2019verifai} introduced a comprehensive validation tool to find falsifying examples via semantic scenario parameters that is usable with external simulators. \cite{dreossi2019compositional} proposed a compositional falsification framework that isolates feature spaces for the non-ML and ML sub-systems to generate counter-examples separately. \cite{ghosh2018verifying} uses signal temporal logic for formalized safety specification and automated falsification of ML-based systems with graph-based bayesian optimization.

The above approaches are effective for efficient falsification but do not offer generative models for massive parallel simulations. Within this area, data-driven approaches are typically employed \cite{ding2022survey}, which use traffic databases to directly sample \cite{webb2020waymo} or learn targeted probability distribution models \cite{o2018scalable}. \cite{o2018scalable} explicitly allows the incorporation of \textit{desired outcomes} that dictate the created distributions. \cite{rempe2022generating} uses a model prior in addition to traffic databases to synthesize both knowledge and data in subsequent models. Data-driven methods, while critical for bridging real and virtual domains, have inherently limited potential of exploring \textit{novel} scenarios via intelligent search algorithms, unlike their rule-based counterparts introduced above. Our proposed methodology integrates an efficient optimization routine in a formalized rule-based scenario framework, which yields a generative model rather than singular counter-examples, which can be used to generate a large number of counter-examples and allows convenient storage and reuse of insights achieved by the optimization run.

\section{Problem Statement}\label{sec:problem}
A scenario $\mathbf{S}$ is composed of participating agents $\mathbb{A}$ and actions $\mathbb{U}$, where each action $u \in \mathbb{U}$ is defined for certain agents $\mathbf{a} \in \mathbb{A}$ and controlling parameters $p \in \mathbb{P}$:

\begin{equation}\label{eq:scenario}
	\mathbf{S} := \{\mathbb{A}, \mathbb{U}\}, \:\: \:\: \:\: 
	u := f(\mathbf{a},p).
\end{equation}

For instance, Sec. \ref{sec:application} specifies a cut-in scenario with two vehicles as agents and cut-in maneuver as action. The parameters such as action trigger times, relative initial positions and velocities control the action exection and form the basis of scenario variation. A simulation model $M$ maps the scenario agents to sub-models and derives initial states $\underline{s}_0$ and input trajectories $U(t)$ as dictated by scenario actions and parameters, such that

\begin{align} 
	\underline{s}(t) &= M(\underline{s}_0, U(t), t),\\
	\mathbf{T} &= \{\underline{s}(t_0),...,\underline{s}(t_N)\}, \:\: t_0 \leq t \leq t_N
\end{align}

where $\underline{s}(t)$ are system states, and $\mathbf{T}$ is a trace containing all state trajectories. An \textit{outcome specification} is defined as a formula $\phi$ of predicates over $\mathbf{T}$, such that ($\vDash$ indicates \textit{is true for})

\begin{equation}\label{eq:cost_func}
	\phi \vDash \mathbf{T} \Leftrightarrow \xi(\mathbf{T})  = 0.
\end{equation}

$\xi$ is a positive real-valued cost function over the trace. The objective is to find a distribution $\mathbb{P}_\theta$ over the scenario parameter space $\mathbb{P}$ such that the resulting simulation and its subsequent trace has a high probability to minimize $\xi$,

\begin{equation}\label{eq:opt_func}
	\underset{\theta}{\mathrm{argmax}} \:\: p_\theta(\xi=0).
\end{equation}

\section{Background}\label{sec:Background}
This section presents a theoretical overview of the methods used in Sec. \ref{sec:Framework} to tackle the objective in \eqref{eq:opt_func}.

\subsection{Signal Temporal Logic}\label{sec:stl}
Signal temporal logic (STL) \cite{maler2004monitoring} is used within the formal verification community for monitoring continuous temporal signals of complex systems and is used to express the outcome specifications in \eqref{eq:cost_func}. A specification (formula) is expressed as a combination of multiple predicates:

\begin{equation}
	\phi := true \:|\: \pi \:|\: \lnot \:\phi \:|\: \phi \land \psi \:|\: \phi \: \mathcal{U}_{[a,b]} \psi ,
\end{equation}

where $\psi$ is another formula and $\mathcal{U}_{[a,b]}$ an until operator over the time interval $[a,b]$. A predicate $\pi$ is defined as a real function at a single time-point of the trace. STL allows formal and intuitive outcome specifications with complex boolean elements such as implication, if-and-only-if, as well as timed qualifiers such as \textit{eventually} ($\mathbf{F}$) or \textit{always} ($\mathbf{G}$) true:

\begin{align}
\mathbf{F}_{[a,b]} \phi &= true \:\: \mathcal{U}_{[a,b]} \:\:\phi \\
&= \exists t' \in [t+a,t+b] \:\: such \:that\:\: \phi \vDash t',\\
\mathbf{G}_{[a,b]} \phi &= \lnot (\mathbf{F}_{[a,b]} \lnot \phi) \\
&= \forall t' \in [t+a,t+b]\:\: such \:that\:\: \phi \vDash t'.
\end{align}

E.g. Sec. \ref{sec:application} uses the time-to-collision based predicate $(1 - ttc > 0)$ to define a specification over the whole trace: $\mathbf{F}_{[0,T]} (1 - ttc > 0)$. The real value of predicates $\pi$ affords a robustness metric $\rho$ to the specification in addition to truth-false inferences. E.g. the robustness of the  $\mathbf{F}_{[0,T]} (1 - ttc > 0)$ is the maximum value of $(1 - ttc)$ throughout the simulation. Robustness for compound specifications can be calculated automatically:

\begin{align}\label{eq:quant_stl}
	\rho^{\phi \land \psi}(t) &:= \mathrm{min}(\rho^\phi(t),\rho^\psi(t)),\\
	\rho^{\phi \: \mathcal{U}_{[a,b]} \: \psi}(t) &:= 
		\underset{\tau \in t + [a,b]}{\mathrm{max}} \left(
				\mathrm{min}\left(
				\rho^\psi(\tau),\underset{s \in [t,\tau]}{\mathrm{min}} \rho^\phi(s)\right)
													\right),
\end{align}

where $\rho^\phi$ indicates robustness of specification $\phi$. Robustness metric impart both qualitative and quantitative semantics to specifications and open the door for mathematical algorithms to automatically search for falsifying traces.

\subsection{Bayesian Optimization}
Bayesian optimization (BO) is known for its sample-efficiency and flexibility in optimizing black-box functions. BO maintains a surrogate function as belief over the black-box function and updates it via adaptive measurements. Typical choice for the surrogate function is a Gaussian Process (GP), but other methods such as random forests are also used \cite{8957442}.

\subsubsection{Gaussian Processes}
Following the formalism in \cite{williams2006gaussian}, GPs model a prior function $ f(x) \sim \mathcal{GP}(m(x),k(x,x'))$ as a joint gaussian distribution over the continuous function values. $m(x)$ is the prior mean and $k(x,x')$ is a kernel that defines covariance between any two points and encodes prior assumptions on the target function. We use the commonly used Matern kernel which provides explicit parameters to control the smoothness of the fitted function. The GP prior is conditioned on measurements $y(\mathbf{x})$ to create a posterior belief, also a GP. For a zero prior mean and non-noisy measurements (since simulations deliver ground-truth data), the posterior GP is

\begin{align}\label{eq:GP}
	f^*|x^*,\mathbf{X},y(\mathbf{x}) &\sim \mathcal{N}(\mu(x^*),\Sigma(x^*,\mathbf{x})) \\
	\mu(x^*) &= k(x^*,\mathbf{x}) K_{\mathbf{x}\mathbf{x}}^{-1} y(\mathbf{x}), \\
	\Sigma(x^*,\mathbf{x}) &= k(x^*,x^*) - k(x^*,\mathbf{x})K_{\mathbf{x}\mathbf{x}}^{-1}k(\mathbf{x},x^*),
\end{align}
where the covariance $\Sigma(x^*,\mathbf{x})$ is between the new points $x^*$ and previously measured $\mathbf{x}$ and $K_{\mathbf{x}\mathbf{x}}$ is the kernel matrix whose entries $K_{\mathbf{x}\mathbf{x}}(i,j) = k(x_i,x_j)$ are evaluated between each of the previously measured points $\mathbf{x}$.

\subsubsection{Acquisition Function} Given a surrogate prior GP, BO optimizes an acquisition function iteratively to choose the next candidate $x^*$ to compute a surrogate posterior. Among the variety of acquisition functions available, we use the Thompson Sampling (TS) \cite{thompson1933likelihood} method which chooses the candidate $x^*$ by

\begin{equation}\label{eq:acq_func}
	x^* = \underset{x \in D}{\mathrm{argmin}} \:\: f^*,
\end{equation}

where $f^*$ is sampled from the surrogate prior. As it is a distribution, samples $f^*$ from the same prior vary from one-another, thus implicitly imparting both exploration and exploitation to \eqref{eq:acq_func}. The posterior calculated with the candidate $x^*$ and its measurement $y(x^*)$ is used as prior for the next iteration, and the process continues until $f^*_{min}$ converges.

\subsection{Gaussian Mixture Models}
Gaussian mixture model are a weighted sum of $N$ gaussian distributions 

\begin{equation}\label{eq:GMM}
	p(x | \theta) = \sum_{i=1}^{N} w_i \mathcal{N}(x; \mu_i,\Sigma_i),
\end{equation}

commonly used as an unsupervised learning technique. The algorithm learns a number of normal distributions based on clusters of given data. Typically, an expectation maximization algorithm is employed to learn the maximum likelihood estimates of hyper-parameters $\theta$ (means $\mu_i$, covariance $\sigma_i$) \cite{bishop2006pattern}. We employ bayesian gaussian mixture models from the python scikit-learn library \cite{scikit-learn} which also learns the optimal number of distributions to fit on the data.

\begin{figure}
	\centering
	\includegraphics[width=0.47\textwidth]{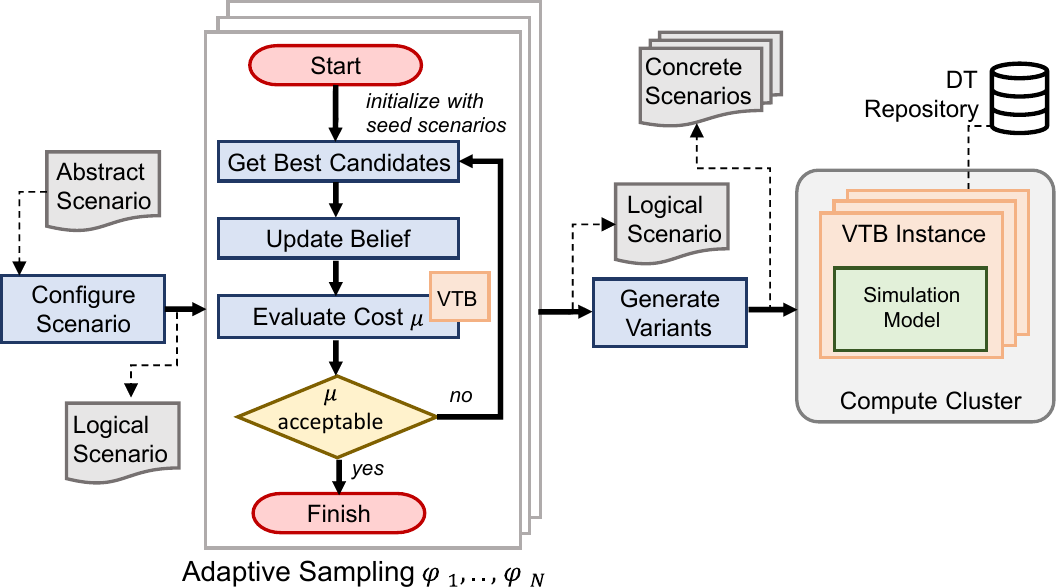}%
	\caption{Framework for scenario-based cluster simulations}
	\label{fig:framework}
\end{figure}

\section{Framework}\label{sec:Framework}

This section presents the framework to solve \eqref{eq:opt_func}, illustrated in Fig. \ref{fig:framework}. The overall workflow can be reiterated as such: given a scenario template and its parameter space, the desired outcomes of scenario simulations are specified as formal specifications. An optimizer learns a distribution over the parameter space via successive simulations so that sampling from said distribution has high likelihood of conforming to given specifications, i.e. delivering desired simulation outcomes. The overall formalism is roughly derived from \cite{9687075} with emphasis on the scenario variation aspect.

\subsubsection{Scenarios}
\begin{figure}
	\centering
	\includegraphics[width=0.37\textwidth]{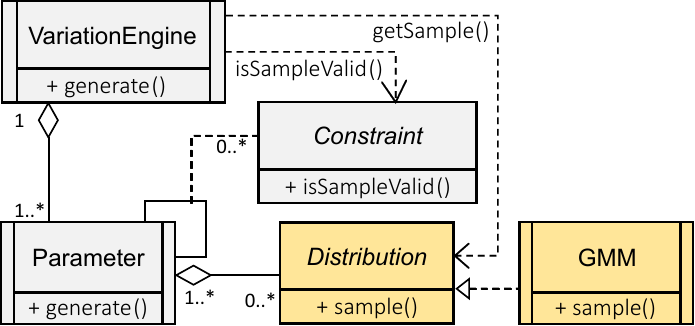}%
	\caption{Meta-model for logical scenario based on \cite{MaqboolR22}}
	\label{fig:log_scn}
\end{figure}

Scenarios specify the actors and events in a human-readable manner independent of the simulation framework. We specify the \textit{static} scenario (road layout etc.) with ASAM OpenDRIVE \cite{Opendrive} format and the \textit{dynamic} scenario (actor behaviour) with OpenSCENARIO \cite{OpenScenario}. Our framework additionally differentiates between the \textit{abstract}, \textit{logical} and \textit{concrete} scenarios. The abstract scenario, referring both the static and dynamic scenario, is complete with respect to actors and events but defines no values for the scenario parameters. The logical scenario allows human designers to specify a scenario parameter space using a formalized meta-model illustrated in Fig. \ref {fig:log_scn} from \cite{MaqboolR22}. Given parameter ranges and inter-parameter relations specified per the \textit{Constraint} and probability distributions per the \textit{Distribution} template, the Variation Engine use Markov-Chain Monte-Carlo methods to efficiently sample the resulting parameter space. This paper focuses on the case where the designer desires certain simulation outcomes but does not know the corresponding parameter distributions or constraints beforehand. These outcomes are specified within the logical scenario as-are, the subsequent adaptive sampling algorithm derives the optimal parameter distribution and saves it as an instance of the \textit{Distribution} template (see the yellow highlighted boxes in Fig. \ref{fig:log_scn}). The Variation Engine can then sample the parameter space to generate concrete scenarios. Concrete scenarios are complete in all aspects, and can be converted to simulation models for the virtual test beds (VTB).

\subsubsection{Outcome Specification}
An outcome specification is defined with the STL formalism introduced in Sec. \ref{sec:stl} over one or more predicates - real-valued functions that can be evaluated over a simulation trace. For each concrete scenario and its subsequent simulation trace, STL formalism allows the evaluation of a robustness value, whose sign indicates conformity and value indicates the extent of conformity. We additionally use a cost metric $\xi$ that affords control over the subsequent optimization algorithms:

\begin{equation}\label{eq:cost_stl}
	\xi(\mathbf{T}) =
	\begin{cases}
		-\rho ^ {\phi_\mathbf{T}} ,& \text{if } \rho ^ {\phi_\mathbf{T}} < 0\\
		0,              & \text{otherwise}.
	\end{cases}
\end{equation}  
The cost metric offers mere convenience and is irrelevant for the optimization routine. \eqref{eq:cost_stl} uses it to invert robustness for minimization and to assert that all scenarios conforming to the specification are equally relevant. Since scenario distributions rather than singular examples are the goal, this encourage the adaptive sampling algorithm to fit on a diverse range of conforming scenarios rather than the most robust scenario.

\subsubsection{Adaptive Sampling}
Illustrated in Fig. \ref{fig:framework}, bayesian optimization (BO) initializes with rasterized parameter samples and their evaluated costs $(p_0,\xi_0)$ to fit a Gaussian Process (GP) $f$ as a surrogate function as in \eqref{eq:GP}. Since $\xi$ must be evaluated on VTB, it is an expensive process, and the idea is not just to find the minimum $\xi$, but to estimate an $f$ that forms a reasonable belief on $\xi$ near minima. This is further encouraged by a batch variant of Thompson Sampling (TS), wherein $N$ surrogate $f$ are optimized simultaneously. In each successive iteration, BO optimizes the TS acquisition function in \eqref{eq:acq_func} to chose the next best candidate parameters. The candidates and their costs $(p_t^*,\xi_t^*)$ thus create a new posterior GP as in \eqref{eq:GP}. 

The converged surrogate forms the basis to infer scenarios likely to yield the desired outcome, as parameters with low cost on a sufficiently-fit surrogate are likely to have lower costs on the VTB as well. A large number of parameters corresponding to minimum cost range of the surrogate are therefore fit on a gaussian mixture model in \eqref{eq:GMM}. The GMM is presented as $\mathbb{P}_\theta$ the solution to \eqref{eq:opt_func}, and serves as an intuitive and lightweight description for the scenario distribution. It is saved within the logical scenario as an instance of the \textit{Distribution} template via a set of hyper-parameter, which can also be tuned to control the variance of concrete scenarios.

\subsubsection*{Example}
\begin{figure}[]
	\centering
	\subcaptionbox{BO surrogate estimation\label{fig:bo_exa}}{\includegraphics[width=0.24\textwidth]{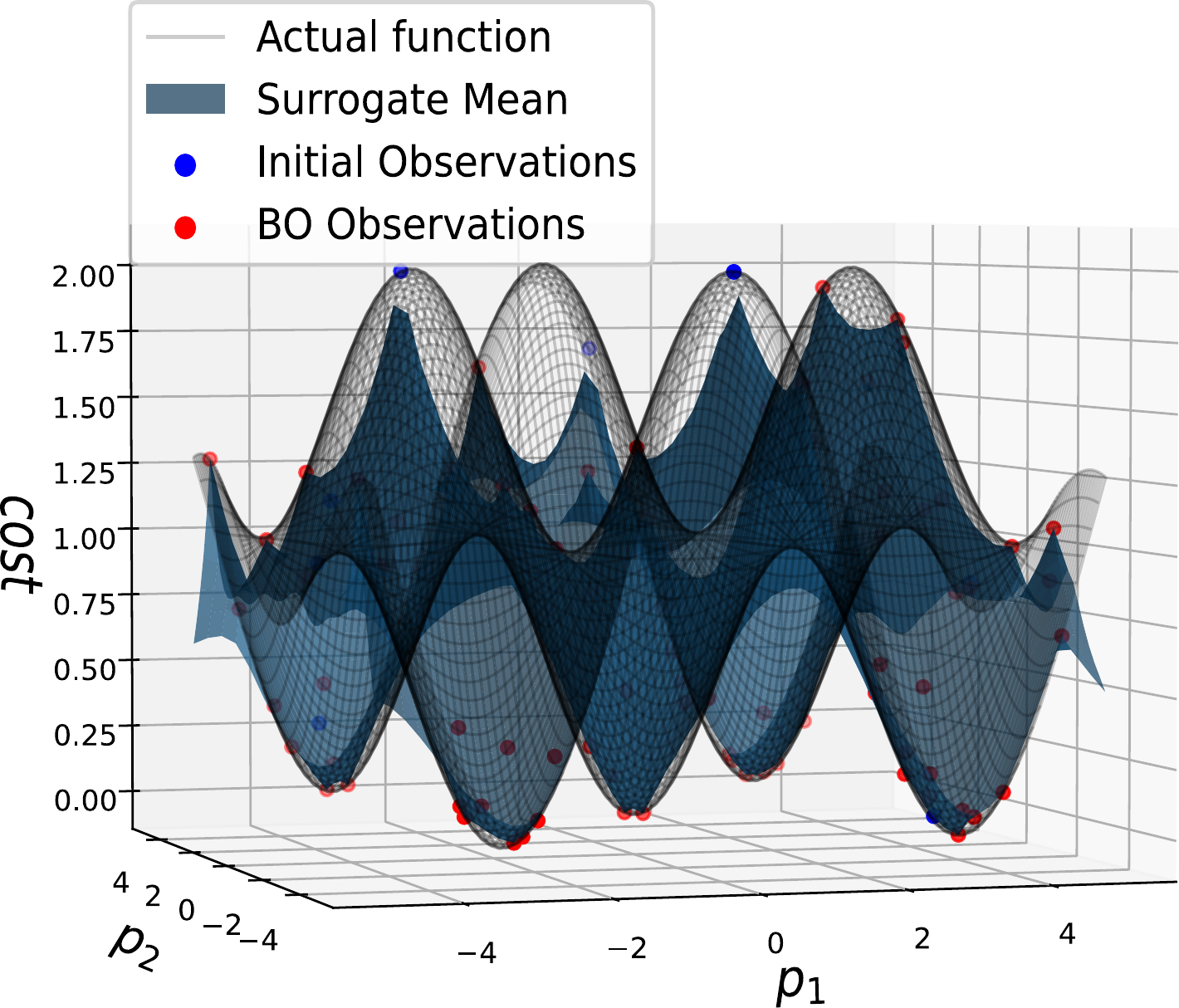}}%
	\hfill
	\subcaptionbox{Samples from fitted GMM\label{fig:bo_exb}}{\includegraphics[width=0.24\textwidth]{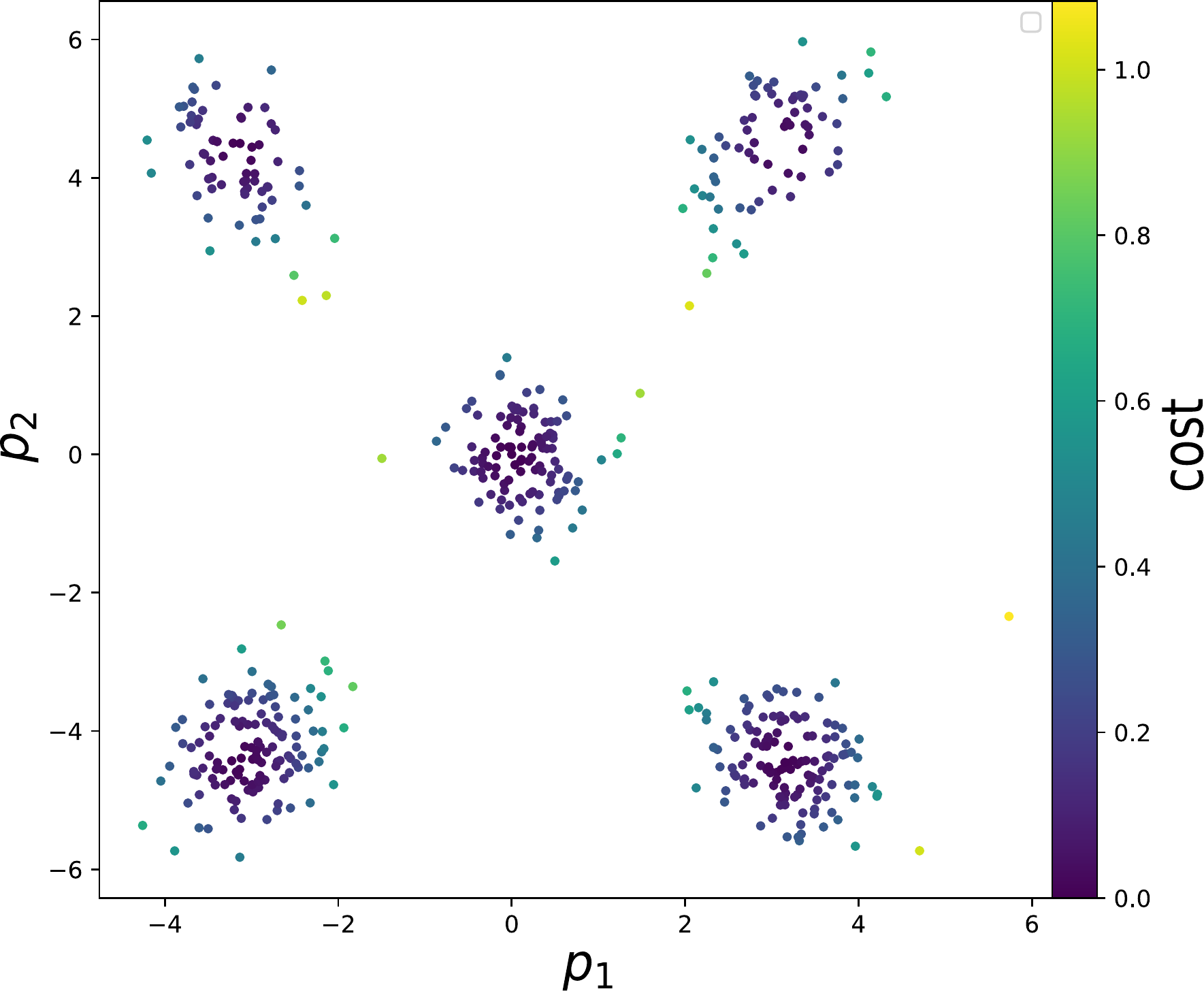}}%
	\caption{Adaptive sampling example on the Griewank function}
	\label{fig:bo_ex}
\end{figure}

An example for the adaptive sampling method is illustrated in Fig. \ref{fig:bo_ex} for a two-dimensional Griewank function \cite{griewank1981generalized}. Fig. \ref{fig:bo_exa} shows that the surrogate function regresses well on the original function around the minima even with existence of multiple minima. This is relevant since all parameter regions corresponding to an outcome are typically desired. The experiment was carried out with an initial draw of 11 samples, and converged in 8 iterations, each with a batch size of 5. The GMM was fitted to parameter values corresponding to the surrogate cost range $[0-0.25]$. Samples from the GMM and their actual function evaluation is illustrated in Fig. \ref{fig:bo_exb}.

\begin{figure*}
	\includegraphics[width=\textwidth]{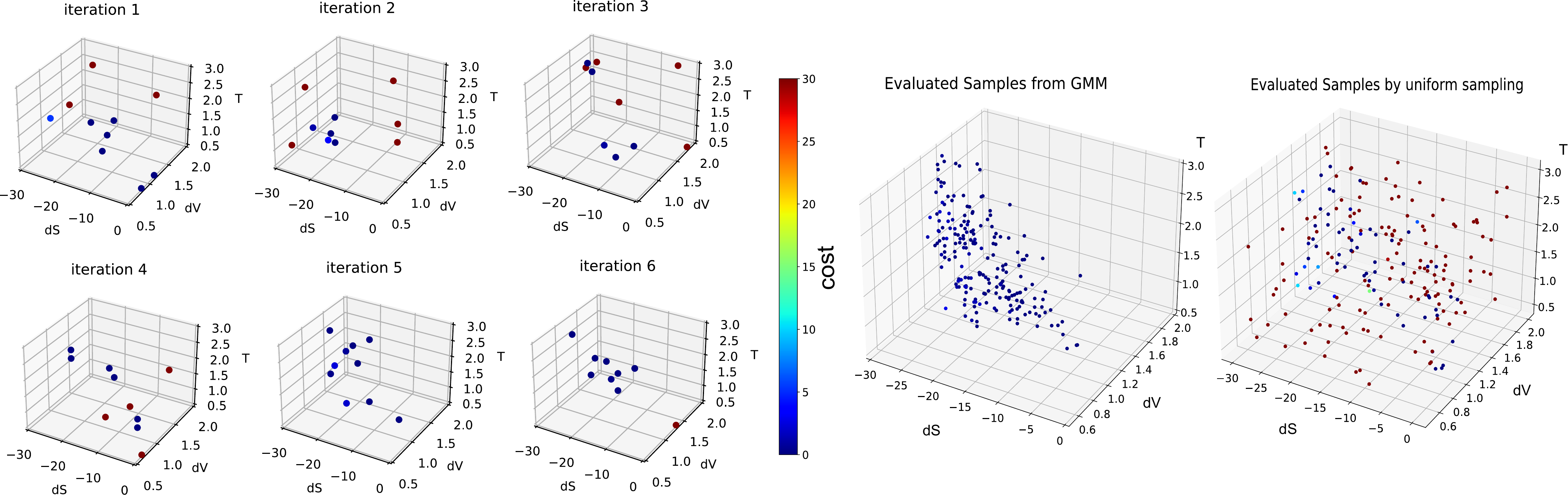}
	\caption{(Left) Bayesian optimization iterations for \textit{cut-in-from-left} scenario, initialized with a rasterized grid (not shown). (Right) Samples from fitted GMM vs uniform samples evaluated on the VTB}
	\label{fig:bo_cutin}
\end{figure*}

\subsubsection{Simulation on VTB Clusters}
Concrete scenarios, sampled from GMM and any other constraints, are imported in VTB instances running on a compute cluster. VTB serves as a cross-domain platform offering modular simulation algorithms (e.g. rigid body dynamics, realistic sensor simulation). For each concrete scenario, a VTB instance constructs a simulation model by mapping actors to digital twins from a central repository, and actions to input trajectories and initial states of the digital twins. We use the framework developed by \cite{9687075} within the VEROSIM platform \cite{rossmann2012control}. Each concrete scenario and subsequent VTB instance are fully independent and self-contained, constituting a freely scalable ``embarassingly parallel" problem. The simulation traces are logged in an SQLite data-bank \cite{atorf2015flexible} accessible throughout the framework cluster.

\section{Transfer in Application}\label{sec:application}
The presented application is from a highway Traffic Jam Chauffeur (TJC) research project, where a machine-learning based TJC module must be comprehensively validated against its operational scenarios. We present one of the scenarios, \textit{cut-in-from-left}, wherein a test vehicle \textit{Vehicle\_1} behind the ego vehicle cuts in its lane from left, and the goal is to find a distribution of critically dangerous scenarios that the ego vehicle may encounter.

\makeatletter
\renewcommand{\ALG@name}{Scenario}
\makeatother

\begin{algorithm}
	\caption{Abstract Scenario: Cut-In-From-Left}\label{scn:abs_scn}
	\footnotesize
	\begin{algorithmic}[1]
		\renewcommand{\algorithmicrequire}{\textbf{Parameters:}}
		\REQUIRE $dS, dV, T$
		\renewcommand{\algorithmicrequire}{\textbf{Init:}}
		\renewcommand{\algorithmicensure}{\textbf{Story:}}
		\REQUIRE 
		\STATE \textbf{Actor} ego  \\ \quad pos $\gets 1000 m$, vel $\gets 16.667 ms^{-1}$, lane $\gets 0$
		\STATE \textbf{Actor}: vehicle\_1 \\ \quad rel\_pos $\gets dS$, rel\_vel $\gets dV$, rel\_lane $\gets -1$
		\ENSURE
		\renewcommand{\algorithmicensure}{\textbf{\quad Action:}}
		\ENSURE Cut-In
		\STATE \quad Actors  $\gets$ vehicle\_1
		\STATE \quad Trigger\_Time $\gets T$
	\end{algorithmic} 
\end{algorithm}

The abstract scenario is defined with OpenSCENARIO formalism and summarized in Scenario \ref{scn:abs_scn}. The initial relative position and velocity of \textit{vehicle\_1}, and trigger time for the cut-in maneuver are undefined parameters, whose corresponding ranges are described in the Logical Scenario \ref{scn:log_scn}, which also states an appropriate outcome specification for \textit{critical scenarios}: the time-to-collision between two vehicles must be below $1s$ at-least once during the simulation. The parameter distribution to achieve high conformity to the specification is specified with a gaussian mixture model (GMM) with unknown hyper-parameters $\theta$.

\begin{figure}
	\centering
	\includegraphics[width=0.37\textwidth]{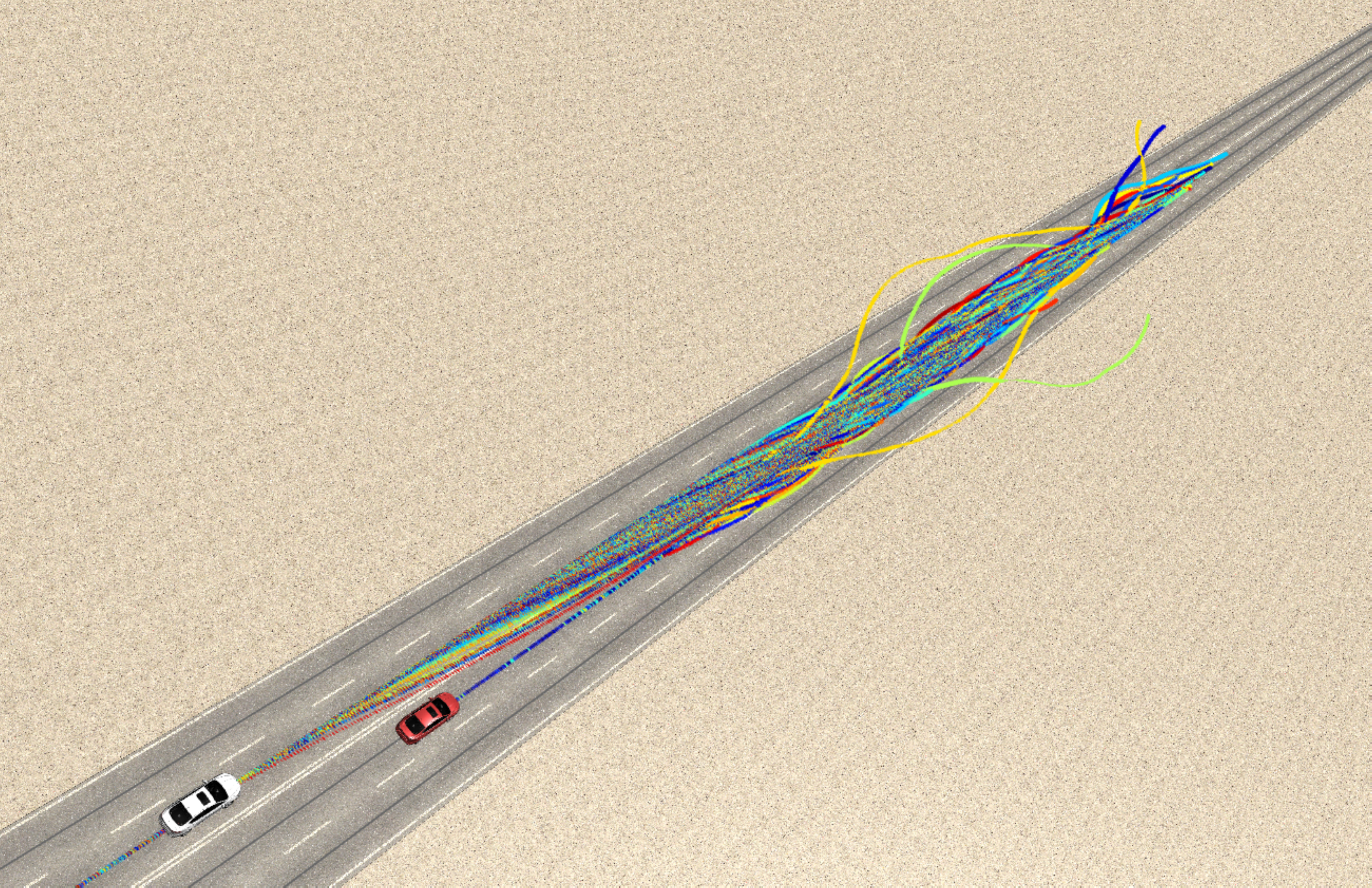}%
	\caption{Vehicle tracks for 200 samples from GMM for $Spec1$.}
	\label{fig:frame_tracks}
\end{figure}

\begin{algorithm}
	\caption{Logical Scenario: Cut-In-From-Left}\label{scn:log_scn}
	\footnotesize
	\begin{algorithmic}[1]
		\renewcommand{\algorithmicrequire}{\textbf{Parameters:}}
		\REQUIRE 
		\STATE  $dS \gets$ range $[-30,0]$ 
		\STATE $dV \gets$ range $[0.5,2.0]$ 
		\STATE $T \gets$ range $[0.5,3.0]$
		\renewcommand{\algorithmicrequire}{\textbf{Outcome Specs:}}
		\REQUIRE 
		\STATE $ Spec1: \mathbf{F} (1 - ttc) > 0$
		\STATE \quad \textbf{Distribution} : GMM $\gets \theta$
	\end{algorithmic} 
\end{algorithm}

The adaptive sampling methodology creates a an initial set of concrete scenarios with 64 rasterized samples of the parameter space. For each concrete scenario, a VTB instance delivers a time-to-collision trace calculated at each simulation time-step via projections of the position and velocity vectors of both vehicles. An STL parser evaluates the outcome specification $Spec 1$ to calculate the cost of each concrete scenario. The bayesian optimization (BO) routine then iteratively suggests new samples to evaluate on the VTB while maintaining a surrogate Guassian Process (GP). Fig. \ref{fig:bo_cutin} (left) illustrates BO for 6 iterations each with a batch size of 10, suggesting samples in optimal yet diverse regions. The converged surrogate GP is used to infer parameter candidates with cost (time-to-collision) less than 1, which are then fitted to a GMM to derive $\theta$. Fig. \ref{fig:bo_cutin} (center) depicts 200 samples from the learned $\theta$ that were evaluated on the VTB. The high percentage of evaluations conforming to $Spec 1$ (zero cost) is evident as compared to 200 uniform samples evaluated on the VTB (Fig. \ref{fig:bo_cutin} right). The VTB evaluations on GMM samples are also visualized interactively using methodologies from \cite{atorf2018interactive} in Fig. \ref{fig:cutin_examples} and \ref{fig:frame_tracks}. Fig. \ref{fig:frame_tracks} visualizes the vehicle tracks for all 200 simulations, whereas Fig. \ref{fig:cutin_examples} shows selected examples in detail.

\section{Conclusion}
This paper proposed generative models to efficiently test relevant use-cases for on-cluster massive parallel simulations. The authors proposed a formal framework based on formal specifications, human-readable scenarios and modular virtual test beds to set up the optimization and simulation tool chain, and presented a bayesian optimization based adaptive sampling algorithm to learn the optimal generative models. The framework was tested on an example scenario and compared with uniform simulations, where the presented method showed substantially better performance.

\bibliographystyle{IEEEtran}
\bibliography{references}

\end{document}